\documentclass[11pt,a4paper]{article}
\usepackage[hyperref]{emnlp-ijcnlp-2019}
\usepackage{times}
\usepackage{latexsym}
\usepackage{booktabs}
\usepackage{url}
\usepackage{graphicx}
\usepackage{amsmath}
\usepackage{amssymb}
\usepackage{xcolor}
\usepackage{bm}
\usepackage[linesnumbered]{algorithm2e}
\usepackage{algorithmic}
\usepackage{textcomp}
\usepackage{caption}
\usepackage{subcaption}

\definecolor{spanish}{RGB}{0, 112, 184} 

\usepackage{color}
\usepackage[textsize=scriptsize]{todonotes}
\definecolor{blue}{RGB}{0, 93, 170}
\definecolor{darkgreen}{RGB}{0, 102, 0}

\aclfinalcopy 

\title{Leveraging Dependency Forest for Neural Medical Relation Extraction}

\author{Linfeng Song$^{1,2}$, Yue Zhang$^{3,4}$\thanks{~~Yue Zhang is the corresponding author}, Daniel Gildea$^1$, Mo Yu$^5$, Zhiguo Wang$^6$ \and Jinsong Su$^7$ \\
  $^1$University of Rochester, Rochester, NY, USA \\
  $^2$Tencent AI Lab, Bellevue, WA, USA \\
  $^3$Institute of Advanced Technology, Westlake Institute for Advanced Study \\
  $^4$School of Engineering, Westlake Universtiy \\
  $^5$IBM T.J. Watson Research, Yorktown Heights, NY, USA \\
  $^6$Amazon AWS, New York, NY, USA \\
  $^7$Xiamen University, Xiamen, China \\
  }

\date{}

\begin{document}
\maketitle
\begin{abstract}
  Medical relation extraction discovers relations between entity mentions in text, such as research articles.
  For this task, dependency syntax has been recognized as a crucial source of features.
  Yet in the medical domain, 1-best parse trees suffer from relatively low accuracies, diminishing their usefulness.
  We investigate a method to alleviate this problem by utilizing dependency forests.
  Forests contain many possible decisions and therefore have higher recall but more noise compared with 1-best outputs.
  A graph neural network is used to represent the forests, automatically distinguishing the useful syntactic information from parsing noise.
  Results on two biomedical benchmarks show that our method outperforms the standard tree-based methods, giving the state-of-the-art results in the literature.
\end{abstract}

\section{Introduction}

\begin{figure}
    \centering
    \includegraphics[width=0.99\linewidth]{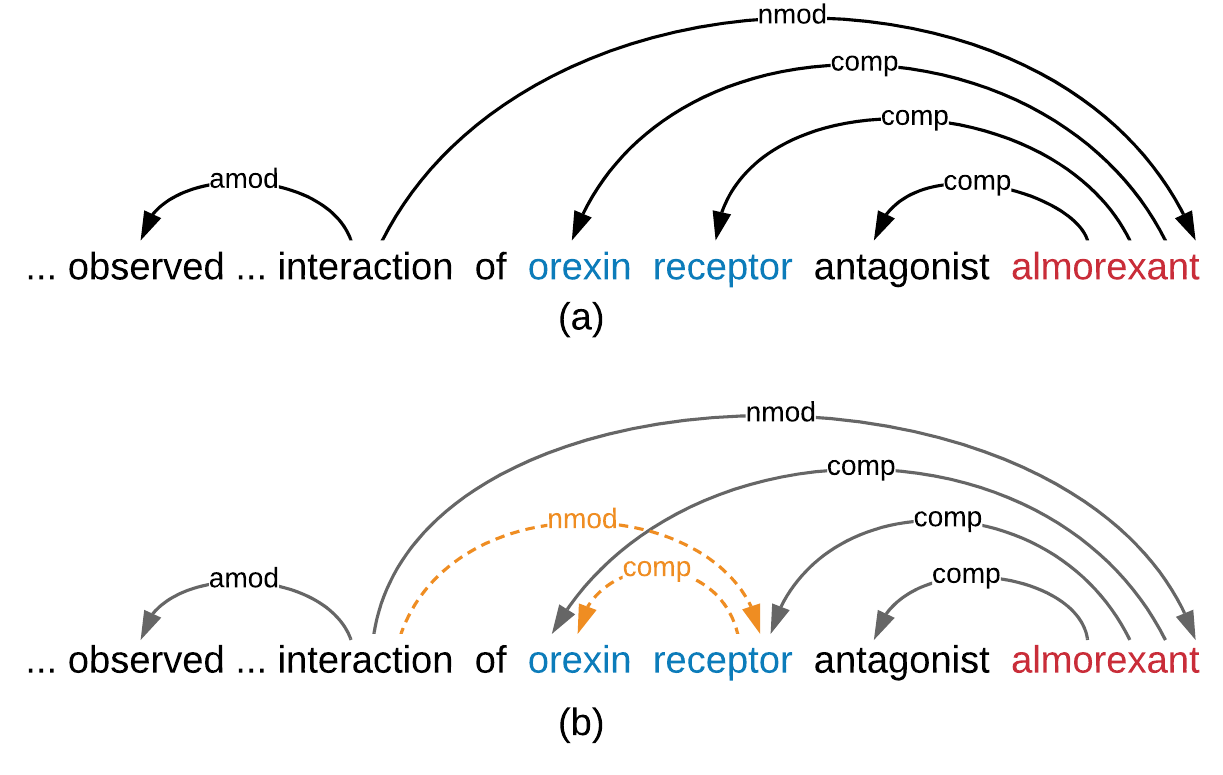}
    \caption{(a) 1-best dependency tree and (b) dependency forest for a medical-domain sentence, where edge label ``comp'' represents ``compound''. Associated mentions are in different colors. Some irrelevant words and edges are omitted for simplicity.}
    \vspace{-1.0em}
    \label{fig:example}
\end{figure}

The sheer amount of medical articles and their rapid growth prevent researchers from receiving comprehensive literature knowledge by direct reading.
This can hamper both medical research and clinical diagnosis.
NLP techniques have been used for automating the knowledge extraction process from the medical literature \citep{friedman2001genies,yu2003extracting,hirschman2005overview,xu2010medex,sondhi2010shallow,abacha2011automatic}.
Along this line of work, a long-standing task is relation extraction, which mines factual knowledge from free text by labeling relations between entity mentions.
As shown in Figure \ref{fig:example}, the sub-clause ``previously observed cytochrome P450 3A4 ( CYP3A4 ) interaction of the dual \textcolor{spanish}{orexin receptor} antagonist \textcolor{red}{almorexant}'' contains two entities, namely ``\textcolor{spanish}{orexin receptor}'' and ``\textcolor{red}{almorexant}''.
There is an ``adversary'' relation between these two entities, denoted as``CPR:6''.

Previous work has shown that dependency syntax is important for guiding relation extraction \citep{culotta-sorensen-2004-dependency,bunescu-mooney-2005-shortest,liu-etal-2015-dependency,gormley2015improved,xu-etal-2015-semantic,xu-etal-2015-classifying,miwa-bansal-2016-end,zhang-etal-2018-graph}, especially in biological and medical domains \cite{quirk-poon:2017:EACLlong,TACL1028,song2018nary}.
Compared with sequential surface-level structures, such as POS tags, dependency trees help to model word-to-word relations more easily by drawing direct connections between distant words that are syntactically correlated.
Take the phrase ``effect on the medicine'' for example; ``effect'' and ``medicine'' are directly connected in a dependency tree, regardless of how many modifiers are added in between.

Dependency parsing has achieved an accuracy over 96\% in the news domain \citep{liu-zhang-2017-order,kitaev-klein-2018-constituency}.
However, for the medical literature domain, parsing accuracies can drop significantly \citep{lease2005parsing,mcclosky2008self,sagae2008evaluating,candito2011word}.
This can lead to severe error propagation in downstream relation extraction tasks, offsetting much of the benefit that relation extraction models can obtain by exploiting dependency trees as a source of external features.

We address the low-accuracy issue in biomedical dependency parsing by considering dependency forests as external features.
Instead of 1-best trees, dependency forests 
consist of dependency arcs and labels that a parser is relatively confident about, therefore having better recall of gold-standard arcs by offering more candidate choices with noise.
Our main idea is to let a relation extraction system learn automatically from a forest which arcs are the most relevant through end-task training, rather than relying solely on the decisions of a noisy syntactic parser.
To this end, a graph neural network is used for encoding a forest, which in turn provides features for relation extraction.
Back-propagation passes loss gradients from the relation extraction layer to the graph encoder, so that the more relevant edges can be chosen automatically for better relation extraction.

Results on BioCreative VI ChemProt (CPR) \citep{krallinger2017overview} and a recent dataset focused on phenotype-gene relations (PGR) \citep{sousa2019silver} 
show that our method outperforms a strong baseline that uses 1-best dependency trees as features, giving the state-of-the-art accuracies in the literature. 
To our knowledge, we are the first to study dependency forests for medical information extraction, showing their advantages over 1-best tree structures.
Our code is available at 
\url{http://github.com/freesunshine0316/dep-forest-re}.

\section{Related work}

\subparagraph{Syntactic forests}
There have been previous studies leveraging constituent forests for machine translation \citep{mi-etal-2008-forest,ma-etal-2018-forest,zaremoodi-haffari-2018-incorporating}, sentiment analysis \citep{le-zuidema-2015-forest} and text generation \citep{lu-ng-2011-probabilistic}.
However, the usefulness of dependency forests is relatively rarely studied, with one exception being
\citet{tu-etal-2010-dependency}, who use dependency forests to enhance long-range word-to-word dependencies for statistical machine translation.
To our knowledge, we are the first to study the usefulness of dependency forests for relation extraction under a strong neural framework.

\vspace{0.5em}
\textbf{Graph neural network}~~
Graph neural networks (GNNs) have been successful in encoding dependency trees for downstream tasks, such as semantic role labeling \citep{marcheggiani-titov-2017-encoding}, semantic parsing \citep{xu-etal-2018-exploiting}, machine translation \citep{song2019semantic,bastings-etal-2017-graph}, relation extraction \citep{song2018nary} and sentence ordering \citep{yin2019graph}.
In particular, \citet{song2018nary} showed that GNNs are more effective than DAG networks \citep{TACL1028} for modeling syntactic trees in relation extraction, which cause loss of important structural information.
We are the first to exploit GNNs for encoding search spaces in the form of dependency forests.

\section{Task}

Formally, the input to our task is a sentence $\boldsymbol{s}= w_1, w_2, \dots, w_N$, where $N$ is the number of words in the sentence and $w_i$ represents the $i$-th input word.
$\boldsymbol{s}$ is annotated with boundary information ($\xi_1:\xi_2$ and $\zeta_1:\zeta_2$) of target entity mentions ($\xi$ and $\zeta$).
We focus on the classic binary relation extraction setting \citep{quirk-poon:2017:EACLlong}, where the number of associated mentions is two.
The output is a relation from a predefined relation set $\boldsymbol{R}=(r_1, \dots, r_M, \text{None})$, where ``None'' means that no relation holds for the entities.

Two steps are taken for predicting the correct relation given an input sentence.
First, a dependency parser is used to label the syntactic structure of the input.
Here our baseline system takes the standard approach, using the 1-best parser output tree $\boldsymbol{D}_T$ as features.
In contrast, our proposed model uses the most confident parser forest $\boldsymbol{D}_F$ as features.
Given $\boldsymbol{D}_T$ or $\boldsymbol{D}_F$, the second step is to encode both $s$ and $\boldsymbol{D}_T/\boldsymbol{D}_F$ using a neural network, before making a prediction.

We make use of the same graph neural network encoder structure to represent dependency syntax information for both the baseline and our model.
In particular, a graph recurrent neural network architecture \citep{beck-etal-2018-graph,song-etal-2018-graph,zhang-etal-2018-sentence} is used, which has been shown effective in encoding graph structures \citep{song2019semantic}, giving competitive results with alternative graph networks such as graph convolutional neural networks \citep{marcheggiani-titov-2017-encoding,bastings-etal-2017-graph}.

\section{Baseline: \textsc{DepTree}}
\label{sec:baseline}

\begin{figure}
    \centering
    \includegraphics[width=0.9\linewidth]{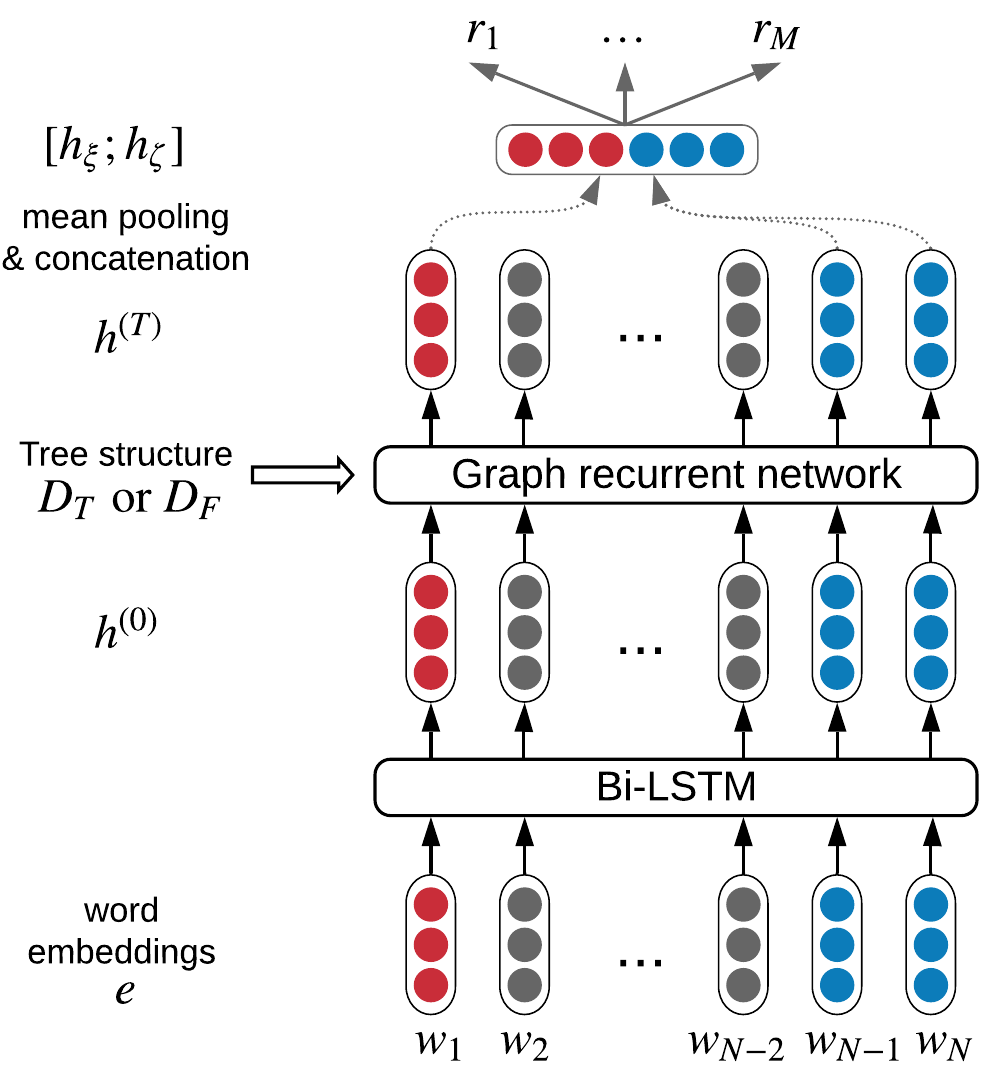}
    \caption{Framework of our baseline and model.}
    \label{fig:model}
\end{figure}

As shown in Figure \ref{fig:model}, our baseline model stacks a bidirectional LSTM layer to encode an input sentence $w_1, \dots, w_N$ with a graph recurrent network (GRN) to encode a 1-best dependency tree, which extracts features from the sentence and the dependency tree $\boldsymbol{D}_T$, respectively.
Similar model frameworks have shown highly competitive performances in previous relation extraction studies \citep{TACL1028,song2018nary}.

\subsection{Bi-LSTM layer}

Given the input sentence $w_1, w_2, \dots, w_N$, we represent each word with its embedding to generate a sequence of embeddings $\boldsymbol{e}_1, \boldsymbol{e}_2, \dots, \boldsymbol{e}_N$.
A Bi-LSTM layer is used to encode the sentences:
\begin{align} \label{eq:bi_lstm_1}
\overleftarrow{\boldsymbol{h}}_{i}^{(0)} &= \textrm{LSTM}_l(\overleftarrow{\boldsymbol{h}}_{i+1}^{(0)}, \boldsymbol{e}_{i}) \\ \label{eq:bi_lstm_2}
\overrightarrow{\boldsymbol{h}}_{i}^{(0)} &= \textrm{LSTM}_r(\overrightarrow{\boldsymbol{h}}_{i-1}^{(0)}, \boldsymbol{e}_{i}) \text{,}
\end{align}
where the state of each word $w_i$ is generated by concatenating the states of both directions:
\begin{equation} \label{eq:lstm_outputs}
    \boldsymbol{h}_{i}^{(0)} = [\overleftarrow{\boldsymbol{h}}_{i}^{(0)}; \overrightarrow{\boldsymbol{h}}_{i}^{(0)}]
\end{equation}

\subsection{GRN layer}

A 1-best dependency tree can be represented as a directed graph $\boldsymbol{D}_T=\langle\boldsymbol{V},\boldsymbol{E}\rangle$, where $\boldsymbol{V}$ includes all words $w_1, w_2, \dots, w_N$ and $\boldsymbol{E}=\{(w_j,l,w_i)\}|_{w_j\in V, w_i\in V}$ represents all dependency edges \citep{marcheggiani-titov-2017-encoding}.
Each triple $(w_j,l,w_i)$ corresponds to a dependency edge, where $w_j$ modifies $w_i$ with an arc label $l$.
Each word $w_i$ is associated with a hidden state that is initialized with the Bi-LSTM output $\boldsymbol{h}_{i}^{(0)}$.
The state representation of the entire tree consists of all word states:
\begin{equation}
    \boldsymbol{h}^{(0)} = \{\boldsymbol{h}_{i}^{(0)}\}_{w_i\in V}
\end{equation}

In order to capture non-local interactions between words, the GRN layer adopts a message passing framework that performs iterative information exchange between directly connected words.
As a result, each word state is updated by absorbing larger contextual information through the message passing process, and a sequence of state transitions $\boldsymbol{h}^{(0)},\boldsymbol{h}^{(1)},\dots$ is generated for the entire tree.
The final state $\boldsymbol{h}^{(T)}=\text{GRN}(\boldsymbol{h}^{(0)}, T)$, where $T$ is a hyperparameter representing the number of state transitions.

\vspace{0.5em}
\textbf{Message passing}~~
The message passing framework takes two main steps within each iteration: message calculation and state update. 
Take $w_i$ and iteration $t$ as the example.
In the first step, separate messages $\boldsymbol{m}^{\uparrow}_i$ and $\boldsymbol{m}^{\downarrow}_i$ are calculated by summing up the messages of its children and parent in the dependency tree, respectively:
\begin{align} \label{eq:sum_1}
    \boldsymbol{m}^{\uparrow}_i &= \sum_{(w_j, l, w_i)\in \boldsymbol{E}_{(\cdot,\cdot,i)}} [\boldsymbol{h}_{j}^{(t-1)}; \boldsymbol{e}_{l}] \\ \label{eq:sum_2}
    \boldsymbol{m}^{\downarrow}_i &= \sum_{(w_i, l, w_k)\in \boldsymbol{E}_{(i,\cdot,\cdot)}} [\boldsymbol{h}_{k}^{(t-1)}; \boldsymbol{e}_{l_{rev}}] \text{,}
\end{align}
where $\boldsymbol{E}_{(\cdot,\cdot,i)}$ and $\boldsymbol{E}_{(i,\cdot,\cdot)}$ represent all edges with a head word $w_i$ and a modifier word $w_i$, respectively, and $\boldsymbol{e}_{l_{rev}}$ represents the embedding of label $l_{rev}$, the reverse version of original label $l$ (such as ``amod-rev'' is the reverse version of ``amod'').
The message from a child or a parent is obtained by simply concatenating its hidden state with the corresponding edge label embedding.

In the second step, GRN uses standard gated operations of LSTM \citep{hochreiter1997long} to update hidden state $\boldsymbol{h}_{i}^{(t-1)}$ with the previously integrated message.
In particular, a cell $\boldsymbol{c}_{i}^{(t)}$ is taken to record memory for $\boldsymbol{h}_{i}^{(t)}$; an input gate $\boldsymbol{i}_{i}^{(t)}$, an output gate $\boldsymbol{o}_{i}^{(t)}$ and a forget gate $\boldsymbol{f}_{i}^{(t)}$ are used to control information flow from the inputs and to the output $\boldsymbol{h}_{i}^{(t)}$:
\begin{equation}
\begin{split}
    \boldsymbol{i}_{i}^{(t)} &= \sigma (\boldsymbol{W}^{\uparrow}_1 \boldsymbol{m}^{\uparrow}_i + \boldsymbol{W}^{\downarrow}_1 \boldsymbol{m}^{\downarrow}_i+\boldsymbol{b}_1) \\
    \boldsymbol{o}_{i}^{(t)} &= \sigma (\boldsymbol{W}^{\uparrow}_2 \boldsymbol{m}^{\uparrow}_i + \boldsymbol{W}^{\downarrow}_2 \boldsymbol{m}^{\downarrow}_i+\boldsymbol{b}_2) \\
    \boldsymbol{f}_{i}^{(t)} &= \sigma (\boldsymbol{W}^{\uparrow}_3 \boldsymbol{m}^{\uparrow}_i + \boldsymbol{W}^{\downarrow}_3 \boldsymbol{m}^{\downarrow}_i+\boldsymbol{b}_3) \\
    \boldsymbol{u}_{i}^{(t)} &= \tanh (\boldsymbol{W}^{\uparrow}_4 \boldsymbol{m}^{\uparrow}_i + \boldsymbol{W}^{\downarrow}_4 \boldsymbol{m}^{\downarrow}_i+\boldsymbol{b}_4) \\
    \boldsymbol{c}_{i}^{(t)} &= \boldsymbol{f}_{i}^{(t)} \odot \boldsymbol{c}_{i}^{(t-1)} + \boldsymbol{i}_{i}^{(t)} \odot \boldsymbol{u}_{i}^{(t)} \\
    \boldsymbol{h}_{i}^{(t)} &= \boldsymbol{o}_{i}^{(t)} \odot \tanh(\boldsymbol{c}_{i}^{(t)}) \text{,}
\end{split}
\end{equation}
where $\boldsymbol{W}^{\uparrow}_x$, $\boldsymbol{W}^{\downarrow}_x$, and $\boldsymbol{b}_x$ ($x \in \{1,2,3,4\}$) are model parameters, and $\boldsymbol{c}_{i}^{(0)}$ is initialized as a vector of zeros.

The same process repeats for $T$ iterations.
Starting from $\boldsymbol{h}^{(0)}$ of the Bi-LSTM layer, increasingly more informed hidden states $\boldsymbol{h}^{(t)}$ are obtained as the iteration increases, and $\boldsymbol{h}^{(T)}$ is used as the final representation of each word.

\subsection{Relation prediction}

Given $\boldsymbol{h}^{(T)}$ of the GRN encoding, we calculate the representation vector of the two related entity mentions $\xi$ and $\zeta$ (such as ``\textcolor{red}{almorexant}'' and ``\textcolor{spanish}{orexin receptor}'' in Figure \ref{fig:example}) with mean pooling:
\begin{align}
    \boldsymbol{h}_{\xi} &= f_{\text{mean}}(\boldsymbol{h}_{\xi_1:\xi_2}^{(T)}) \\
    \boldsymbol{h}_{\zeta} &= f_{\text{mean}}(\boldsymbol{h}_{\zeta_1:\zeta_2}^{(T)})
\end{align}
where $\xi_1:\xi_2$ and $\zeta_1:\zeta_2$ represent the span of $\xi$ and $\zeta$, respectively, and $f_{\text{mean}}$ is the mean-pooling function.

Finally, the representations of both mentions are concatenated to be the input of a logistic regression classifier:
\begin{equation}
    \boldsymbol{y} = \text{softmax}(\boldsymbol{W}_5 [\boldsymbol{h}_{\xi};\boldsymbol{h}_{\zeta}] + \boldsymbol{b}_5) \text{,}
\end{equation}
where $\boldsymbol{W}_5$ and $\boldsymbol{b}_5$ are model parameters.

\section{Model}

In this section, we first discuss how to generate high-quality dependency forests, before showing how to adapt GRN to consider the parser probability of each dependency edge.

\subsection{Forest generation}
\label{sec:algo}

Given a dependency parser, generating dependency forests with high recall and low noise is a non-trivial problem.
On the one hand, keeping the whole search space gives 100\% recall, but introduces maximum noise.
On the other hand, using the 1-best dependency tree can result in low recall given an imperfect parser.
We investigate two algorithms to generate high-quality forests by judging ``quality'' from different perspectives: one focusing on arcs, and the other focusing on trees.

\subparagraph{\textsc{Edgewise}}
This algorithm focuses on the local relation of each individual edge and uses parser probabilities as confidence scores to assess edge qualities.
Starting from the whole parser search space, it keeps all the edges with scores greater than a threshold $\gamma$.
The time complexity is $O(N^2)$, where $N$ represents the sentence length.\footnote{More accurately, it is $O(N^2L)$ and $L$ s a \emph{constant} factor, denoting the number of distinct dependency labels. We omit it for simplicity.}

\subparagraph{\textsc{KBestEisner}}
This algorithm extends the Eisner algorithm \citep{eisner-1996-three} with cube pruning \citep{huang2005better} for finding $K$ highest-scored tree structures.
The Eisner algorithm is a standard method for decoding 1-best trees for graph-based dependency parsing.
Based on bottom-up dynamic programming, it stores the 1-best subtree for each span and takes $O(N^3)$ time complexity for decoding a sentence of $N$ words.

\textsc{KBestEisner} keeps a sorted list of $K$-best hypotheses for each span.
Cube pruning \citep{huang2005better} is adopted to generate the $K$-best list for each larger span from the $K$-best lists of its sub-spans.
After the bottom-up decoding, we merge the final $K$-bests by combining identical dependency edges to make the forest.
As a result, \textsc{KBestEisner} takes $O(N^3K\log K)$ time.

\vspace{0.5em}
\textbf{Discussions}~~
\textsc{Edgewise} is much simpler and faster than \textsc{KBestEisner}.
Compared with the $O(N^3K\log K)$ time complexity of \textsc{KBestEisner}, \textsc{Edgewise} only takes $O(N^2)$ running time, and each step (storing an edge) runs faster than \textsc{KBestEisner} (making a new hypothesis by combining two from sub-spans).
Besides, the forests of \textsc{Edgewise} can be denser and provide richer information
than those from \textsc{KBestEisner}.
This is because \textsc{KBestEisner} only merges $K$ trees, where many edges are shared among them.
Also, $K$ cannot be set to a large number (such as 100), because that will cause a dramatic increase of running time.

Compared with \textsc{KBestEisner}, \textsc{Edgewise} suffers from two potential problems.
First, \textsc{Edgewise} does not guarantee to produce a 1-best tree in a generated forest, as it makes decisions by considering the individual edges.
Second, it does not guarantee to generate spanning forests, which can happen when the threshold $\gamma$ is high.
On the other hand, no previous work has shown that the information from the whole tree is crucial for relation extraction.
In fact, many previous studies use only the dependency path between the target entity mentions \citep{bunescu-mooney-2005-shortest,airola2008all,chowdhury-etal-2011-study,gormley2015improved,mehryary-etal-2016-deep}.
We study the effectiveness of both algorithms in our experiments.

\subsection{GRN encoding with parser confidence}

As illustrated by Figure \ref{fig:example}(b), our dependency forests are directed graphs that can be consumed by GRN without any structural changes.
For fair comparison, we use the same model as the baseline to encode sentences and forests. 
Thus our model uses \textbf{the same number of parameters} as our baseline taking 1-best trees.

Since forests contain more than one tree, it is intuitive to consider parser confidence scores for potentially better feature extraction.
To this end, we slightly adjust the GRN encoding process without introducing additional parameters.
In particular, we enhance the original message sum function (Equations \ref{eq:sum_1} and \ref{eq:sum_2}) by applying the edge probabilities in calculating weighted message sums:
\begin{align}
    \boldsymbol{m}^{\uparrow}_i &= \sum_{\epsilon\in \boldsymbol{E}_{(\cdot,\cdot,i)}} p_{\epsilon} [\boldsymbol{h}_{j}^{(t-1)}; \boldsymbol{e}_{l}] \\
    \boldsymbol{m}^{\downarrow}_i &= \sum_{\epsilon\in \boldsymbol{E}_{(i,\cdot,\cdot)}} p_{\epsilon} [\boldsymbol{h}_{k}^{(t-1)}; \boldsymbol{e}_{l_{rev}}] \text{,}
\end{align}
where $\epsilon$ (instead of a triple) is used to represent an edge for simplicity, and $p_{\epsilon}$ is the parser probability for edge $\epsilon$.
The edge probabilities are \emph{not} adjusted during end-task training.

\section{Training}

\textbf{Relation loss}~~
Given a set of training instances, each containing a sentence $\boldsymbol{s}$ with two target mentions $\xi$ and $\zeta$, and a dependency structure $\boldsymbol{D}$ (tree or forest), we train our models with a cross-entropy loss between the gold-standard relations $r$ and model distribution:
\begin{equation}
l_{R} = -\log p(r|\boldsymbol{s},\xi,\zeta,\boldsymbol{D};\boldsymbol{\theta}) \text{,}
\end{equation}
where $\boldsymbol{\theta}$ represents the model parameters.

\vspace{0.5em}
\textbf{Using additional NER loss}~~
For training on BioCreative VI CPR, we follow previous work \citep{liu2017attention,verga-etal-2018-simultaneously} to take NER loss as additional supervision, though the mention boundaries are known during testing.
\begin{equation}
l_{NER} = -\frac{1}{N}\sum_{n=1}^{N}\log p(t_{n}|\boldsymbol{s},\boldsymbol{D};\boldsymbol{\theta}) \text{,}
\end{equation}
where $t_{n}$ is the gold NE tag of $w_n$ with the ``BIO'' scheme.
Both losses are conditionally independent given the deep features produced by our model, and the final loss for BioCreative VI CPR training is $l=l_{R} + l_{NER}$.

\section{Experiments}

We conduct experiments on two medical benchmarks to test the usefulness of dependency forest.

\subsection{Data}

\subparagraph{BioCreative VI CPR \citep{krallinger2017overview}}
This task\footnote{https://biocreative.bioinformatics.udel.edu/tasks/biocreative-vi/track-5/} focuses on the relations between chemical compounds (such as drugs) and proteins (such as genes).
The full corpus contains 1020, 612 and 800 extracted PubMed\footnote{https://www.ncbi.nlm.nih.gov/pubmed/} abstracts for training, development and testing, respectively.
All abstracts are manually annotated with the boundaries of entity mentions and the relations.
The data provides three types of NEs: ``CHEMICAL'', ``GENE-Y'' and ``GENE-N'', and the relation set $\boldsymbol{R}$ contains 5 regular relations (``CPR:3'', ``CPR:4'', ``CPR:5'', ``CPR:6'' and ``CPR:9'') and the ``None'' relation.

For efficient generation of dependency structures, we segment each abstract into sentences, keeping only the sentences that contain at least a chemical mention and a protein mention.
For any sentence containing several chemical mentions or protein mentions, we keep multiple copies of it with each copy having different target mention pairs.
As a result, we only consider the relations of mentions in the same sentence, assigning all cross-sentence chemical-protein pairs as ``None'' relation.
By doing this, we effectively sacrifice cross-sentence relations, which has a negative effect on our systems; but this is necessary for efficient generation of dependency structures since directly parsing a short paragraph is slow and erroneous.\footnote{\citet{TACL1028} describe a solution for cross-sentence cases, which joins different dependency structures by connecting their roots. We leave it for future work.}
In general, we obtain 16,107 training, 10,030 development and 14,269 testing instances, in which around 23\% have regular relations.
The highest recalls for relations on our development and test sets are 92.25 and 92.54, respectively, because of the exclusion of cross-sentence relations in preprocessing.
We report F1 scores of the full test set for a fair comparison, using all gold regular relations to calculate recalls.

\vspace{0.5em}
\textbf{Phenotype-Gene relation (PGR) \cite{sousa2019silver}}~~
This dataset concerns the relations between human phenotypes (such as diseases) with human genes, where the relation set is a \emph{binary} class on whether a phenotype is related to a gene.
It has 18,451 silver training instances and 220 high-quality test instances, with each containing mention boundary annotations.
We separate the first 15\% training instances as our development set.
Unlike \emph{BioCreative VI CPR}, almost every relation of PGR is within a single sentence.

\subsection{Models}

We compare the following models:
\begin{itemize}
    \item \textsc{TextOnly}: It does not take dependency structures and directly uses the Bi-LSTM outputs ($\boldsymbol{h}^{(0)}$ in Eq. \ref{eq:lstm_outputs}) to make predictions.
    \item \textsc{DepTree}: Our baseline using 1-best dependency trees, as shown in Section \ref{sec:baseline}.
    \item \textsc{EdgewisePS} and \textsc{Edgewise}: Our models using the forests generated by our \textsc{Edgewise} algorithm with or without parser scores.
    \item \textsc{KBestEisnerPS} and \textsc{KBestEisner}: Our model using the forests generated by our \textsc{KBestEisner} algorithm with or without parser scores, respectively.
\end{itemize}

\subsection{Settings}

We take a state-of-the-art deep biaffine parser \citep{dozat2016deep}, trained on the Penn Treebank (PTB) \citep{marcus19building} converted to Universal Dependency, to obtain 1-best trees and full search spaces for generating forests.
Using standard PTB data split (02--21 for training, 22 for development and 23 for testing), it gives UAS and LAS scores of 95.7 and 94.6, respectively.

For the other hyper-parameters, word embeddings are
initialized with the 200-dimensional BioASQ vectors\footnote{http://bioasq.lip6.fr/tools/BioASQword2vec/}, pretrained on 10M abstracts of biomedical articles, and are fixed during training.
The dimension of hidden vectors in Bi-LSTM is set to 200, and the number of message passing steps $T$ is set to 2 based on \citet{zhang-etal-2018-graph}.
We use Adam \citep{kingma2014adam}, with a learning rate of 0.001, as the optimizer.
The batch size, coefficient for $l2$ normalization loss and dropout rate are 20, $10^{-8}$ and 0.1, respectively.

\subsection{Analyses of generated forests}

\begin{table}
    \centering
    \begin{tabular}{lccc}
        \toprule
        $\gamma$ & \#Edge/\#Node & LAS & Conn. Ratio(\%) \\
        \midrule
        0.05 & 2.09 & 92.5 & 100.0 \\
        0.1 & 1.57 & 91.2 & 99.5 \\
        0.2 & 1.34 & 90.5 & 94.2 \\
        0.3 & 1.04 & 88.0 & 77.6 \\
        \bottomrule
        \toprule
        $K$ & \#Edge/\#Node & LAS & Conn. Ratio(\%) \\
        \midrule
        1 & 1.00 & 86.4 & 100.0 \\
        2 & 1.03 & 87.3 & 100.0 \\
        5 & 1.09 & 89.1 & 100.0 \\
        10 & 1.14 & 89.8 & 100.0 \\
        \bottomrule
    \end{tabular}
    \caption{Statistics on forests generated with various $\gamma$ (upper half) and $K$ (lower half) on the development set.}
    \label{tab:stat}
\end{table}

Table \ref{tab:stat} demonstrates several characteristics of the generated forests of both the \textsc{Edgewise} and \textsc{KBestEisner} algorithms in Section \ref{sec:algo}, where ``\#Edge/\#Sent'' measures the forest density with the number of edges divided by the sentence length, ``LAS'' represents the  oracle LAS score on 100 biomedical sentences with manually annotated dependency trees, and ``Conn. Ratio (\%)'' shows the percentage of forests where both related entity mentions are connected.

Regarding the forest density, forests produced by \textsc{Edgewise} generally contain more edges than those from \textsc{KBestEisner}.
Due to the combinatorial property of forests, \textsc{Edgewise} can give much more candidate trees (and sub-trees) for the whole sentence (and each sub-span).
This coincides with the fact that the forests generated by \textsc{Edgewise} have higher oracle scores than these generated by \textsc{KBestEisner}.

For connectivity, \textsc{KBestEisner} guarantees to generate spanning forests.
On the other hand, the connectivity ratio for the forests produced by \textsc{Edgewise} drops when increasing the threshold $\gamma$.
We can have more than 94\% being connected with $\gamma\le0.2$.
Later we will show that good end-task performance can still be achieved with the 94\% connectivity ratio.
This indicates that losing connectivity for a small potion of the data may not hurt the overall performance.

\subsection{Development results}
\label{sec:dev}

\begin{figure}
    \centering
     \begin{subfigure}{0.95\linewidth}
         \centering
         \includegraphics[width=\textwidth]{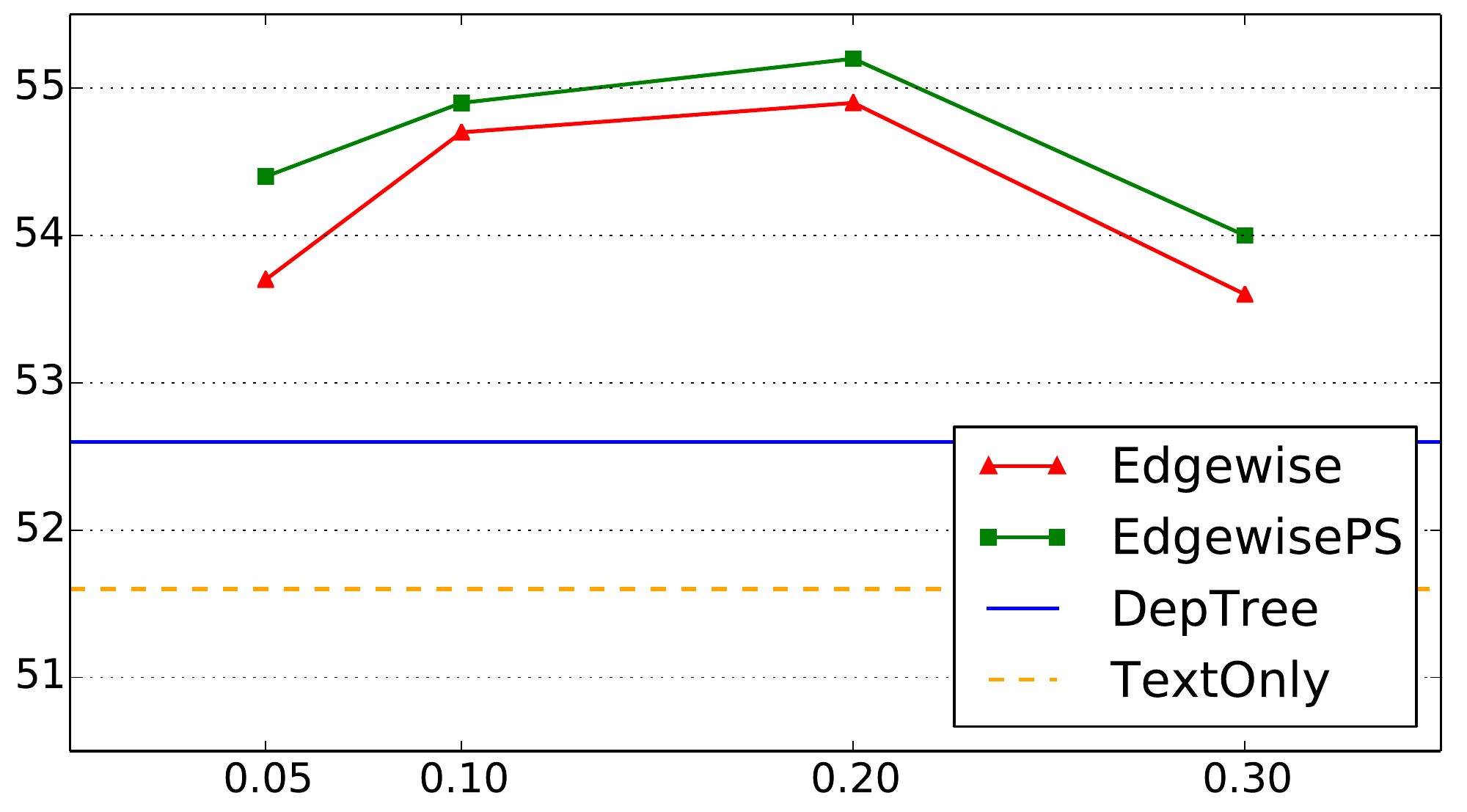}
         \caption{\textsc{Edgewise}}
         \label{fig:dev_cube}
     \end{subfigure}
     \hfill
     \begin{subfigure}[b]{0.95\linewidth}
         \centering
         \includegraphics[width=\textwidth]{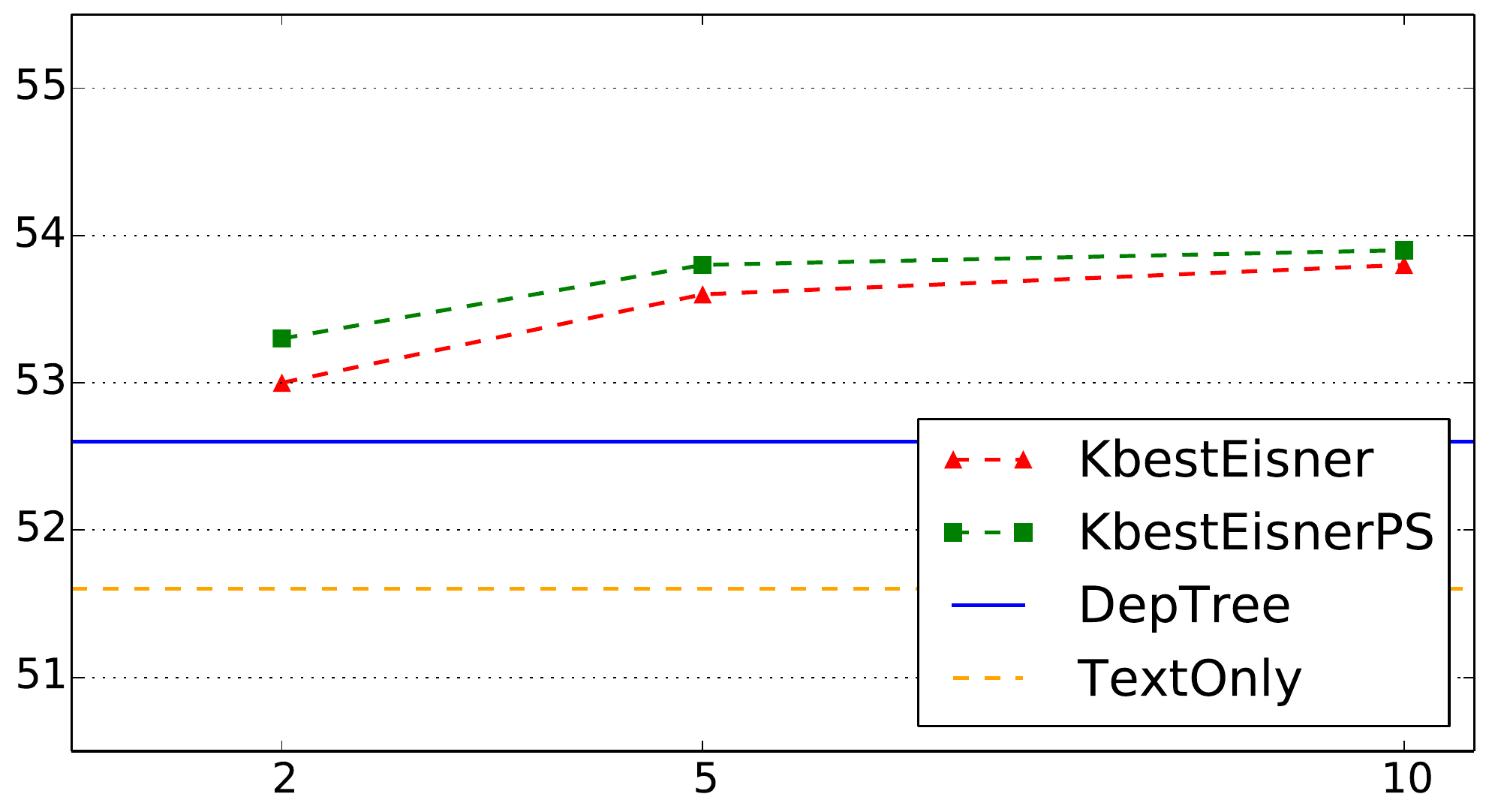}
         \caption{\textsc{KBestEisner}}
         \label{fig:dev_nbest}
     \end{subfigure}
     \vspace{-0.5em}
     \caption{Development results (F1 score) for our forest generation methods.}
     \label{fig:dev}
     \vspace{-1.0em}
\end{figure}

Figure \ref{fig:dev} shows the development experiments for our forest generation algorithms, where both \textsc{Edgewise} and \textsc{KBestEisner} give consistent improvements over \textsc{DepTree} and \textsc{TextOnly}.
Generally, \textsc{Edgewise} gives more improvements than \textsc{KBestEisner}.
The main reason may be that \textsc{Edgewise} generates denser forests, providing richer features.
On the other hand, \textsc{KBestEisner} shows a marginal improvement by increasing $K$ from 5 to 10.
This indicates that only merging 10-best trees may be far from sufficient.
However, using a much larger $K$ (such as 100) is not practical due to dramatically increased computation time.
In particular, the running time of \textsc{KBestEisner} with $K=10$ is already much longer than that of \textsc{Edgewise}.
As a result, \textsc{Edgewise} better serves our goal compared to \textsc{KBestEisner}.
This may sound surprising, as \textsc{Edgewise} does not consider tree-level scores.
It suggests that relation extraction may not require full dependency tree features.
This coincides with previous relation extraction research \citep{bunescu-mooney-2005-shortest,airola2008all}, which utilizes the shortest path connecting the two candidate entities in the dependency tree.

Leveraging parser confidence scores also consistently helps both methods.
It is especially effective for \textsc{Edgewise} when $\gamma=0.05$.
This is likely because the parser confidence scores are useful for distinguishing some erroneous dependency arcs, when noise is large (e.g. when $\gamma$ is too small).
Following the development results, we directly report the performances of \textsc{EdgewisePS} and \textsc{KBestEisnerPS}, setting $\gamma$ and $K$ to 0.2 and 10, respectively, in our remaining experiments.

\subsection{Main results on BioCreative VI CPR}
\label{sec:main_cpr}

\begin{table}
    \centering
    \begin{tabular}{rc}
        \toprule
        Model & F1 score \\
        \midrule
        GRU+Attn \citep{liu2017attention}$\dagger$ & 49.5 \\
        Bran \citep{verga-etal-2018-simultaneously}$\dagger$ & 50.8 \\
        \midrule
        \textsc{TextOnly} & 50.6  \\
        \textsc{DepTree} & 51.4 \\
        \textsc{KBestEisnerPS} & \phantom{**}52.4** \\
        \textsc{EdgewisePS} & \phantom{**}\textbf{53.4}** \\
        \bottomrule
    \end{tabular}
    \caption{Test results of Biocreative VI CPR. $\dagger$ indicates previously reported numbers. ** means significant over \textsc{DepTree} at $p<0.01$ with 1000 bootstrap tests \citep{efron1994introduction}.}
    \label{tab:main_result_cpr}
\end{table}

Table \ref{tab:main_result_cpr} shows the main comparison results on the BioCreative CPR testset, with comparisons to the previous state-of-the-art and our baselines.
\emph{GRU+Attn} \citep{liu2017attention} stacks a self-attention layer on top of GRU \citep{cho-etal-2014-learning} and embedding layers; \emph{Bran} \citep{verga-etal-2018-simultaneously} adopts a biaffine self-attention model to simultaneously extract the relations of all mention pairs.
Both methods use only textual knowledge.

\textsc{TextOnly} gives a performance comparable with \emph{Bran}.
With 1-best dependency trees, our \textsc{DepTree} baseline gives better performances than the previous state of the art.
This confirms the usefulness of dependency structures and the effectiveness of GRN on encoding these structures.
Using dependency forests and parser confidence scores, both \textsc{KBestEisnerPS} and \textsc{EdgewisePS} obtain significantly higher numbers than \textsc{DepTree}.
Consistent with the development experiments, \textsc{EdgewisePS} has a higher testset performance than \textsc{KBestEisnerPS}.

\subsection{Analysis}

\subparagraph{Effectiveness on parsing accuracy}
We have shown in Sections \ref{sec:dev} and \ref{sec:main_cpr} that a dependency parser trained using a domain-general treebank can produce high-quality dependency forests in a target domain (biomedical) for helping relation extraction.
This is based on the assumption of there being a high-quality treebank in a descent scale, which may not be true for low-resource languages.
We simulate this low-resource effect by training our parser in much smaller treebanks of 1K or 5K dependency trees, respectively.
The LAS scores for the resulting parsers on our 100 manually annotated biomedical dependency trees are 79.3 and 84.2, respectively, while the LAS score for the parser trained with the full treebank is 86.4, as shown in Table \ref{tab:stat}.

\begin{figure}
    \centering
    \includegraphics[width=0.95\linewidth]{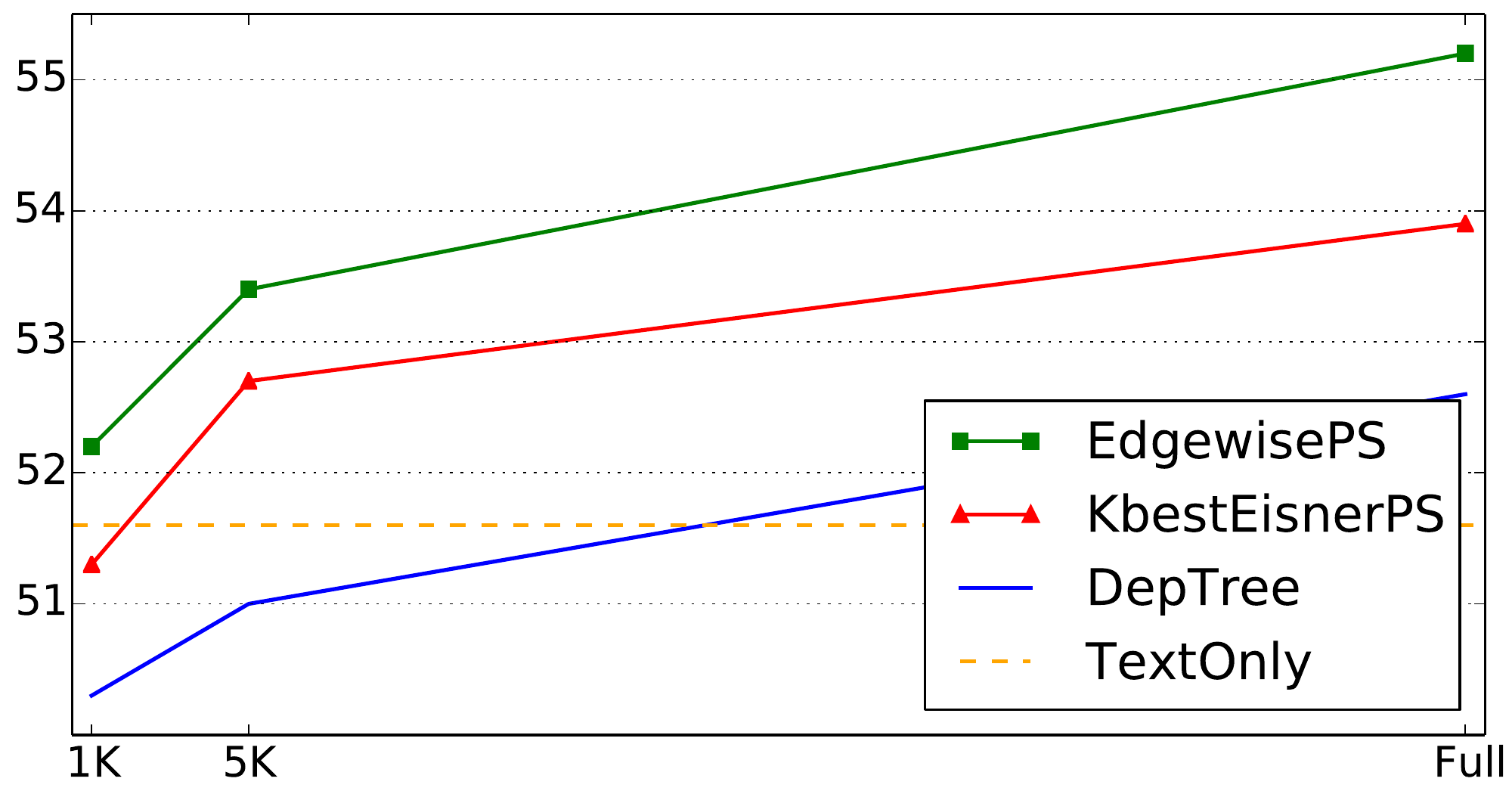}
    \caption{\textsc{Dev} results of BioCreative CPR regarding the dependency parsers trained on different number (1K, 5K or Full) of dependency trees.}
    \label{fig:low_res}
    \vspace{-1.0em}
\end{figure}

Figure \ref{fig:low_res} shows the results on the Biocreative CPR development set, where the performance of \textsc{TextOnly} is 51.6.
\textsc{DepTree} fails to outperform \textsc{TextOnly} when only 1K or 5K dependency trees are available for training our parser.
This is due to the low parsing recall and subsequent noise caused by the weak parsers.
It confirms the previous conclusion that dependency structures are highly influential to the performance of relation extraction.
Both \textsc{EdgewisePS} and \textsc{KBestEisnerPS} are still more effective than \textsc{DepTree}.
In particular, \textsc{KBestEisnerPS} significantly improves \textsc{TextOnly} with 5K dependency trees, and \textsc{EdgewisePS} is helpful even with 1K dependency trees.

\textsc{KBestEisner} shows relatively smaller gaps than \textsc{Edgewise} when only a limited number of dependency trees are available.
This is probably because considering whole-tree quality helps to better eliminate noise.

\begin{figure}
    \centering
    \includegraphics[width=\linewidth]{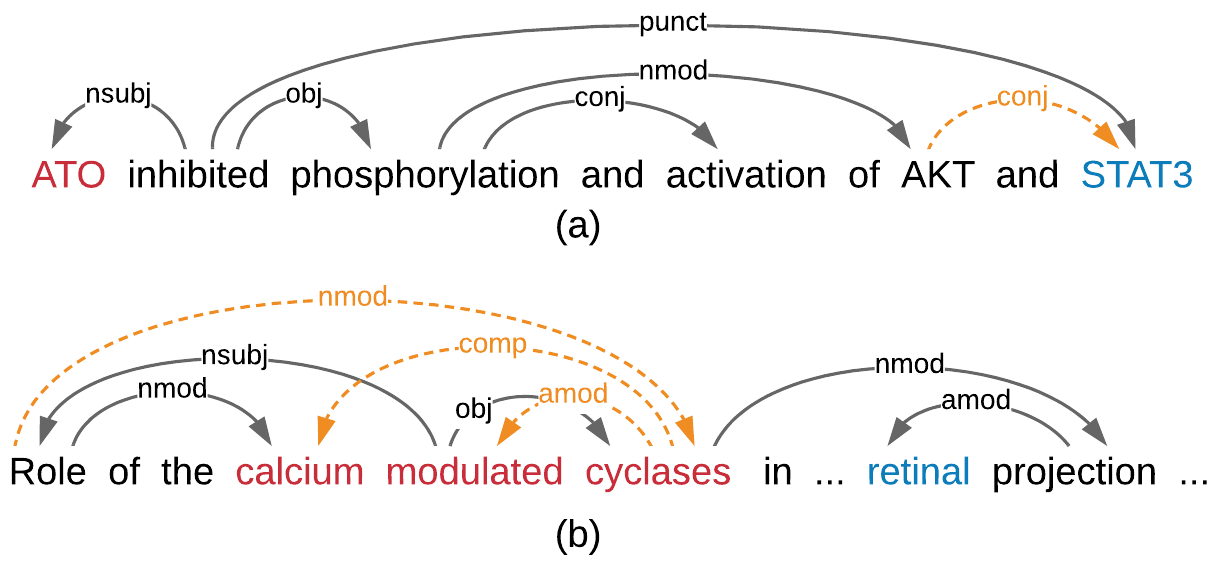}
    \caption{Two representative cases in BioCreative CPR, contrasting 1-best trees and forests, where irrelevant content and arcs are omitted for simplicity.}
    \label{fig:case_study}
\end{figure}

\vspace{0.5em}
\textbf{Case study}~~
Figure \ref{fig:case_study} illustrates two major types of errors in BioCreative CPR, which are caused by inaccurate 1-best dependency trees.
As shown in Figure \ref{fig:case_study}(a), the baseline system mistakenly predicts a ``None'' relation for that instance.
This is mainly because ``\textcolor{spanish}{STAT3}'' is incorrectly linked to the main verb ``inhibited'' with a ``punct'' relation, but it should be linked to ``AKT''.
In contrast, our forest contains the correct relation and with a probability of 0.18.
This is possibly because ``AKT and \textcolor{spanish}{STAT3}'' fits the common pattern of ``A and B'' that conjunct two nouns.

\begin{table}
    \centering
    \begin{tabular}{rc}
        \toprule
        Model & F1 score \\
        \midrule
        BO-LSTM \citep{lamurias2019bo}$\dagger$ & 52.3 \\
        BioBERT \citep{lee2019biobert}$\dagger$ & 67.2 \\
        \midrule
        \textsc{TextOnly} & 76.0 \\
        \textsc{DepTree} & 78.9 \\
        \textsc{KBestEisnerPS} & \phantom{*}83.6* \\
        \textsc{EdgewisePS} & \phantom{**}\textbf{85.7}** \\
        \bottomrule
    \end{tabular}
    \caption{Main results on PGR testest. $\dagger$ denotes previous numbers rounded into 3 significant digits. * and ** indicate significance over \textsc{DepTree} at $p<0.05$ and $p<0.01$ with 1000 bootstrap tests.}
    \label{tab:main_result_pgr}
    \vspace{-1.0em}
\end{table}

Figure \ref{fig:case_study}(b) shows another type of parsing errors that cause end-task mistakes.
In this example, the multi-token mention ``\textcolor{red}{calcium modulated cyclases}'' is incorrectly segmented in the 1-best dependency tree, where ``modulated'' is used as the main verb of the whole sentence, leaving ``cyclases'' and ``calcium'' as the object and the modifier of the subject, respectively.
However, this mention ought to be a noun phrase with ``cyclases'' being the head.
Our forest helps in this case by providing a more reasonable structure (shown as the yellow dashed arcs), where both ``calcium'' and ``modulated'' modify ``cyclases''.
This is likely because ``modulated'' can be interpreted as an adjective in addition to being a verb.
It shows the advantage of keeping multiple candidate syntactic arcs.

\subsection{Main results on PGR}

Table \ref{tab:main_result_pgr} shows the comparison with previous work on the PGR testset, where our models are significantly better than the existing models.
This is likely because the previous models do not utilize all the information from inputs: \emph{BO-LSTM} only takes the words (without arc labels) along the shortest dependency path between the target mentions; the pretrained weights of \emph{BioBERT} are kept constant during training for relation extraction.

With 1-best trees, \textsc{DepTree} is 2.9 points better than \textsc{TextOnly}, confirming the usefulness of dependency structures.
Leveraging dependency forests, both \textsc{KBestEisnerPS} and \textsc{EdgewisePS} significantly outperform \textsc{DepTree} with $p$-values of 0.003 and 0.024, respectively.
This further confirms the usefulness of dependency forests for medical relation extraction.

\subsection{Main results on SemEval-2010 task 8}

\begin{table}
    \centering
    \begin{tabular}{rc}
    \toprule
    Model & F1 score \\
    \midrule
    C-GCN \citep{zhang-etal-2018-graph}$\dagger$ & 84.8 \\
    C-AGGCN \citep{guo-etal-2019-attention}$\dagger$ & 85.7 \\
    \textsc{DepTree} & 84.6 \\
    \midrule
    \textsc{KBestEisnerPS} & 85.8 \\
    \textsc{EdgewisePS} & 86.3 \\
    \bottomrule
    \end{tabular}
    \caption{Main results on SemEval-2010 task 8 testest. $\dagger$ denotes previous numbers.}
    \label{tab:semeval}
    \vspace{-1.0em}
\end{table}

In addition to the biomedical domain, leveraging dependency forests applies to other domains as well.
As shown in Table \ref{tab:semeval}, we conduct a preliminary study on SemEval-2010 task 8 \citep{hendrickx2009semeval}, a widely used benchmark for news-domain relation extraction.
It is a public dataset, containing 10,717 instances (8000 for training and development, 2717 for testing) with 19 relations: 9 directed relations and a special ``Other'' class.
Both \emph{C-GCN} and \emph{C-AGGCN} take a similar network as ours by stacking a graph neural network for encoding trees on top of a Bi-LSTM layer for encoding sentences.

\textsc{DepTree} achieves similar performance as \emph{C-GCN} and is slightly worse than \emph{C-AGGCN}, with one potential reason being that \emph{C-AGGCN} takes more parameters.
Using forests, both \textsc{KBestEisnerPS} and \textsc{EdgewisePS} outperform \textsc{DepTree} with the same number of parameters, and they show comparable and slightly better performances than \emph{C-AGGCN}.
Again, \textsc{EdgewisePS} is better than \textsc{KBestEisnerPS}, showing that the former is a better way for generating forests.

\section{Conclusion}

We have proposed two algorithms for generating high-quality dependency forests for relation extraction, and studied a graph recurrent network for effectively distinguishing useful features from noise in parsing forests.
Experiments on two biomedical relation extraction benchmarks show the superiority of forests versus tree structures, without introducing any additional model parameters.
Our deep analyses indicate that the main advantage comes from alleviating out-of-domain parsing errors.


\paragraph{Acknowledgments}
Research supported by NSF award IIS-1813823.

\bibliography{emnlp-ijcnlp-2019}

\begin{thebibliography}{56}
\expandafter\ifx\csname natexlab\endcsname\relax\def\natexlab#1{#1}\fi

\bibitem[{Abacha and Zweigenbaum(2011)}]{abacha2011automatic}
Asma~Ben Abacha and Pierre Zweigenbaum. 2011.
\newblock Automatic extraction of semantic relations between medical entities:
  a rule based approach.
\newblock \emph{Journal of biomedical semantics}, 2(5).

\bibitem[{Airola et~al.(2008)Airola, Pyysalo, Bj{\"o}rne, Pahikkala, Ginter,
  and Salakoski}]{airola2008all}
Antti Airola, Sampo Pyysalo, Jari Bj{\"o}rne, Tapio Pahikkala, Filip Ginter,
  and Tapio Salakoski. 2008.
\newblock All-paths graph kernel for protein-protein interaction extraction
  with evaluation of cross-corpus learning.
\newblock \emph{BMC bioinformatics}, 9(11):S2.

\bibitem[{Bastings et~al.(2017)Bastings, Titov, Aziz, Marcheggiani, and
  Simaan}]{bastings-etal-2017-graph}
Joost Bastings, Ivan Titov, Wilker Aziz, Diego Marcheggiani, and Khalil Simaan.
  2017.
\newblock Graph convolutional encoders for syntax-aware neural machine
  translation.
\newblock In \emph{Proceedings of the 2017 Conference on Empirical Methods in
  Natural Language Processing}.

\bibitem[{Beck et~al.(2018)Beck, Haffari, and Cohn}]{beck-etal-2018-graph}
Daniel Beck, Gholamreza Haffari, and Trevor Cohn. 2018.
\newblock Graph-to-sequence learning using gated graph neural networks.
\newblock In \emph{Proceedings of the 56th Annual Meeting of the Association
  for Computational Linguistics (Volume 1: Long Papers)}.

\bibitem[{Bunescu and Mooney(2005)}]{bunescu-mooney-2005-shortest}
Razvan Bunescu and Raymond Mooney. 2005.
\newblock A shortest path dependency kernel for relation extraction.
\newblock In \emph{Proceedings of the conference on human language technology
  and empirical methods in natural language processing}.

\bibitem[{Candito et~al.(2011)Candito, Anguiano, and Seddah}]{candito2011word}
Marie Candito, Enrique~Henestroza Anguiano, and Djam{\'e} Seddah. 2011.
\newblock A word clustering approach to domain adaptation: Effective parsing of
  biomedical texts.
\newblock In \emph{Proceedings of the 12th International Conference on Parsing
  Technologies}.

\bibitem[{Cho et~al.(2014)Cho, van Merrienboer, Gulcehre, Bahdanau, Bougares,
  Schwenk, and Bengio}]{cho-etal-2014-learning}
Kyunghyun Cho, Bart van Merrienboer, Caglar Gulcehre, Dzmitry Bahdanau, Fethi
  Bougares, Holger Schwenk, and Yoshua Bengio. 2014.
\newblock Learning phrase representations using {RNN} encoder{--}decoder for
  statistical machine translation.
\newblock In \emph{Proceedings of the 2014 Conference on Empirical Methods in
  Natural Language Processing (EMNLP)}.

\bibitem[{Chowdhury et~al.(2011)Chowdhury, Lavelli, and
  Moschitti}]{chowdhury-etal-2011-study}
Faisal~Md. Chowdhury, Alberto Lavelli, and Alessandro Moschitti. 2011.
\newblock A study on dependency tree kernels for automatic extraction of
  protein-protein interaction.
\newblock In \emph{Proceedings of {B}io{NLP} 2011 Workshop}.

\bibitem[{Culotta and Sorensen(2004)}]{culotta-sorensen-2004-dependency}
Aron Culotta and Jeffrey Sorensen. 2004.
\newblock Dependency tree kernels for relation extraction.
\newblock In \emph{Proceedings of the 42nd Meeting of the Association for
  Computational Linguistics (ACL{'}04), Main Volume}.

\bibitem[{Dozat and Manning(2017)}]{dozat2016deep}
Timothy Dozat and Christopher~D Manning. 2017.
\newblock Deep biaffine attention for neural dependency parsing.
\newblock In \emph{Proceedings of International Conference on Learning
  Representations}.

\bibitem[{Efron and Tibshirani(1994)}]{efron1994introduction}
Bradley Efron and Robert~J Tibshirani. 1994.
\newblock \emph{An introduction to the bootstrap}.
\newblock CRC press.

\bibitem[{Eisner(1996)}]{eisner-1996-three}
Jason~M. Eisner. 1996.
\newblock Three new probabilistic models for dependency parsing: An
  exploration.
\newblock In \emph{Proceedings of the 16th conference on Computational
  linguistics-Volume 1}.

\bibitem[{Friedman et~al.(2001)Friedman, Kra, Yu, Krauthammer, and
  Rzhetsky}]{friedman2001genies}
Carol Friedman, Pauline Kra, Hong Yu, Michael Krauthammer, and Andrey Rzhetsky.
  2001.
\newblock Genies: a natural-language processing system for the extraction of
  molecular pathways from journal articles.
\newblock \emph{Bioinformatics}, 17(1).

\bibitem[{Gormley et~al.(2015)Gormley, Yu, and Dredze}]{gormley2015improved}
Matthew~R Gormley, Mo~Yu, and Mark Dredze. 2015.
\newblock Improved relation extraction with feature-rich compositional
  embedding models.
\newblock In \emph{Proceedings of the 2015 Conference on Empirical Methods in
  Natural Language Processing}, pages 1774--1784.

\bibitem[{Guo et~al.(2019)Guo, Zhang, and Lu}]{guo-etal-2019-attention}
Zhijiang Guo, Yan Zhang, and Wei Lu. 2019.
\newblock Attention guided graph convolutional networks for relation
  extraction.
\newblock In \emph{Proceedings of the 57th Annual Meeting of the Association
  for Computational Linguistics}, pages 241--251.

\bibitem[{Hendrickx et~al.(2009)Hendrickx, Kim, Kozareva, Nakov,
  {\'O}~S{\'e}aghdha, Pad{\'o}, Pennacchiotti, Romano, and
  Szpakowicz}]{hendrickx2009semeval}
Iris Hendrickx, Su~Nam Kim, Zornitsa Kozareva, Preslav Nakov, Diarmuid
  {\'O}~S{\'e}aghdha, Sebastian Pad{\'o}, Marco Pennacchiotti, Lorenza Romano,
  and Stan Szpakowicz. 2009.
\newblock Semeval-2010 task 8: Multi-way classification of semantic relations
  between pairs of nominals.
\newblock In \emph{Proceedings of SemEval}, pages 94--99.

\bibitem[{Hirschman et~al.(2005)Hirschman, Yeh, Blaschke, and
  Valencia}]{hirschman2005overview}
Lynette Hirschman, Alexander Yeh, Christian Blaschke, and Alfonso Valencia.
  2005.
\newblock Overview of biocreative: critical assessment of information
  extraction for biology.

\bibitem[{Hochreiter and Schmidhuber(1997)}]{hochreiter1997long}
Sepp Hochreiter and J{\"u}rgen Schmidhuber. 1997.
\newblock Long short-term memory.
\newblock \emph{Neural computation}, 9(8).

\bibitem[{Huang and Chiang(2005)}]{huang2005better}
Liang Huang and David Chiang. 2005.
\newblock Better k-best parsing.
\newblock In \emph{Proceedings of the Ninth International Workshop on Parsing
  Technology}.

\bibitem[{Kingma and Ba(2014)}]{kingma2014adam}
Diederik~P Kingma and Jimmy Ba. 2014.
\newblock Adam: A method for stochastic optimization.
\newblock \emph{arXiv preprint arXiv:1412.6980}.

\bibitem[{Kitaev and Klein(2018)}]{kitaev-klein-2018-constituency}
Nikita Kitaev and Dan Klein. 2018.
\newblock Constituency parsing with a self-attentive encoder.
\newblock In \emph{Proceedings of the 56th Annual Meeting of the Association
  for Computational Linguistics (Volume 1: Long Papers)}.

\bibitem[{Krallinger et~al.(2017)Krallinger, Rabal, Akhondi
  et~al.}]{krallinger2017overview}
Martin Krallinger, Obdulia Rabal, Saber~A Akhondi, et~al. 2017.
\newblock Overview of the biocreative vi chemical-protein interaction track.
\newblock In \emph{Proceedings of the VI BioCreative challenge evaluation
  workshop}.

\bibitem[{Lamurias et~al.(2019)Lamurias, Sousa, Clarke, and
  Couto}]{lamurias2019bo}
Andre Lamurias, Diana Sousa, Luka~A Clarke, and Francisco~M Couto. 2019.
\newblock {BO-LSTM}: classifying relations via long short-term memory networks
  along biomedical ontologies.
\newblock \emph{BMC bioinformatics}, 20(1):10.

\bibitem[{Le and Zuidema(2015)}]{le-zuidema-2015-forest}
Phong Le and Willem Zuidema. 2015.
\newblock The forest convolutional network: Compositional distributional
  semantics with a neural chart and without binarization.
\newblock In \emph{Proceedings of the 2015 Conference on Empirical Methods in
  Natural Language Processing}.

\bibitem[{Lease and Charniak(2005)}]{lease2005parsing}
Matthew Lease and Eugene Charniak. 2005.
\newblock Parsing biomedical literature.
\newblock In \emph{International Conference on Natural Language Processing}.

\bibitem[{Lee et~al.(2019)Lee, Yoon, Kim, Kim, Kim, So, and
  Kang}]{lee2019biobert}
Jinhyuk Lee, Wonjin Yoon, Sungdong Kim, Donghyeon Kim, Sunkyu Kim, Chan~Ho So,
  and Jaewoo Kang. 2019.
\newblock Biobert: pre-trained biomedical language representation model for
  biomedical text mining.
\newblock \emph{arXiv preprint arXiv:1901.08746}.

\bibitem[{Liu and Zhang(2017)}]{liu-zhang-2017-order}
Jiangming Liu and Yue Zhang. 2017.
\newblock In-order transition-based constituent parsing.
\newblock \emph{Transactions of the Association for Computational Linguistics},
  5.

\bibitem[{Liu et~al.(2017)Liu, Shen, Wang, Rastegar-Mojarad, Elayavilli,
  Chaudhary, and Liu}]{liu2017attention}
Sijia Liu, Feichen Shen, Yanshan Wang, Majid Rastegar-Mojarad,
  Ravikumar~Komandur Elayavilli, Vipin Chaudhary, and Hongfang Liu. 2017.
\newblock Attention-based neural networks for chemical protein relation
  extraction.
\newblock In \emph{Proceedings of the BioCreative VI Workshop}.

\bibitem[{Liu et~al.(2015)Liu, Wei, Li, Ji, Zhou, and
  WANG}]{liu-etal-2015-dependency}
Yang Liu, Furu Wei, Sujian Li, Heng Ji, Ming Zhou, and Houfeng WANG. 2015.
\newblock A dependency-based neural network for relation classification.
\newblock In \emph{Proceedings of the 53rd Annual Meeting of the Association
  for Computational Linguistics and the 7th International Joint Conference on
  Natural Language Processing (Volume 2: Short Papers)}.

\bibitem[{Lu and Ng(2011)}]{lu-ng-2011-probabilistic}
Wei Lu and Hwee~Tou Ng. 2011.
\newblock A probabilistic forest-to-string model for language generation from
  typed lambda calculus expressions.
\newblock In \emph{Proceedings of the Conference on Empirical Methods in
  Natural Language Processing}.

\bibitem[{Ma et~al.(2018)Ma, Tamura, Utiyama, Zhao, and
  Sumita}]{ma-etal-2018-forest}
Chunpeng Ma, Akihiro Tamura, Masao Utiyama, Tiejun Zhao, and Eiichiro Sumita.
  2018.
\newblock Forest-based neural machine translation.
\newblock In \emph{Proceedings of the 56th Annual Meeting of the Association
  for Computational Linguistics (Volume 1: Long Papers)}.

\bibitem[{Marcheggiani and Titov(2017)}]{marcheggiani-titov-2017-encoding}
Diego Marcheggiani and Ivan Titov. 2017.
\newblock Encoding sentences with graph convolutional networks for semantic
  role labeling.
\newblock In \emph{Proceedings of the 2017 Conference on Empirical Methods in
  Natural Language Processing}.

\bibitem[{Marcus and Marcinkiewicz(1993)}]{marcus19building}
Mitchell~P Marcus and Mary~Ann Marcinkiewicz. 1993.
\newblock Building a large annotated corpus of {English}: The {Penn} treebank.
\newblock \emph{Computational Linguistics}, 19(2).

\bibitem[{McClosky and Charniak(2008)}]{mcclosky2008self}
David McClosky and Eugene Charniak. 2008.
\newblock Self-training for biomedical parsing.
\newblock In \emph{Proceedings of the 46th Annual Meeting of the Association
  for Computational Linguistics}.

\bibitem[{Mehryary et~al.(2016)Mehryary, Bj{\"o}rne, Pyysalo, Salakoski, and
  Ginter}]{mehryary-etal-2016-deep}
Farrokh Mehryary, Jari Bj{\"o}rne, Sampo Pyysalo, Tapio Salakoski, and Filip
  Ginter. 2016.
\newblock Deep learning with minimal training data: {T}urku{NLP} entry in the
  {B}io{NLP} shared task 2016.
\newblock In \emph{Proceedings of the 4th {B}io{NLP} Shared Task Workshop}.

\bibitem[{Mi et~al.(2008)Mi, Huang, and Liu}]{mi-etal-2008-forest}
Haitao Mi, Liang Huang, and Qun Liu. 2008.
\newblock Forest-based translation.
\newblock In \emph{Proceedings of ACL-08: HLT}.

\bibitem[{Miwa and Bansal(2016)}]{miwa-bansal-2016-end}
Makoto Miwa and Mohit Bansal. 2016.
\newblock End-to-end relation extraction using lstms on sequences and tree
  structures.
\newblock In \emph{Proceedings of the 54th Annual Meeting of the Association
  for Computational Linguistics (Volume 1: Long Papers)}.

\bibitem[{Peng et~al.(2017)Peng, Poon, Quirk, Toutanova, and Yih}]{TACL1028}
Nanyun Peng, Hoifung Poon, Chris Quirk, Kristina Toutanova, and Wen-tau Yih.
  2017.
\newblock Cross-sentence n-ary relation extraction with graph {LSTMs}.
\newblock \emph{Transactions of the Association for Computational Linguistics},
  5:101--115.

\bibitem[{Quirk and Poon(2017)}]{quirk-poon:2017:EACLlong}
Chris Quirk and Hoifung Poon. 2017.
\newblock Distant supervision for relation extraction beyond the sentence
  boundary.
\newblock In \emph{Proceedings of the 15th Conference of the European Chapter
  of the ACL (EACL-17)}.

\bibitem[{Sagae et~al.(2008)Sagae, Miyao, S{\ae}tre, and
  Tsujii}]{sagae2008evaluating}
Kenji Sagae, Yusuke Miyao, Rune S{\ae}tre, and Jun'ichi Tsujii. 2008.
\newblock Evaluating the effects of treebank size in a practical application
  for parsing.
\newblock In \emph{Software Engineering, Testing, and Quality Assurance for
  Natural Language Processing}.

\bibitem[{Sondhi et~al.(2010)Sondhi, Gupta, Zhai, and
  Hockenmaier}]{sondhi2010shallow}
Parikshit Sondhi, Manish Gupta, ChengXiang Zhai, and Julia Hockenmaier. 2010.
\newblock Shallow information extraction from medical forum data.
\newblock In \emph{Proceedings of the 23rd International Conference on
  Computational Linguistics: Posters}.

\bibitem[{Song et~al.(2019)Song, Gildea, Zhang, Wang, and
  Su}]{song2019semantic}
Linfeng Song, Daniel Gildea, Yue Zhang, Zhiguo Wang, and Jinsong Su. 2019.
\newblock Semantic neural machine translation using amr.
\newblock \emph{Transactions of the Association for Computational Linguistics},
  7:19--31.

\bibitem[{Song et~al.(2018{\natexlab{a}})Song, Zhang, Wang, and
  Gildea}]{song-etal-2018-graph}
Linfeng Song, Yue Zhang, Zhiguo Wang, and Daniel Gildea. 2018{\natexlab{a}}.
\newblock A graph-to-sequence model for amr-to-text generation.
\newblock In \emph{Proceedings of the 56th Annual Meeting of the Association
  for Computational Linguistics (Volume 1: Long Papers)}, pages 1616--1626.

\bibitem[{Song et~al.(2018{\natexlab{b}})Song, Zhang, Wang, and
  Gildea}]{song2018nary}
Linfeng Song, Yue Zhang, Zhiguo Wang, and Daniel Gildea. 2018{\natexlab{b}}.
\newblock N-ary relation extraction using graph-state lstm.
\newblock In \emph{Proceedings of the 2018 Conference on Empirical Methods in
  Natural Language Processing}, pages 2226--2235.

\bibitem[{Sousa et~al.(2019)Sousa, Lam{\'u}rias, and Couto}]{sousa2019silver}
Diana Sousa, Andr{\'e} Lam{\'u}rias, and Francisco~M Couto. 2019.
\newblock A silver standard corpus of human phenotype-gene relations.
\newblock In \emph{Proceedings of the 2019 Conference of the North American
  Chapter of the Association for Computational Linguistics: Human Language
  Technologies}.

\bibitem[{Tu et~al.(2010)Tu, Liu, Hwang, Liu, and
  Lin}]{tu-etal-2010-dependency}
Zhaopeng Tu, Yang Liu, Young-Sook Hwang, Qun Liu, and Shouxun Lin. 2010.
\newblock Dependency forest for statistical machine translation.
\newblock In \emph{Proceedings of the 23rd International Conference on
  Computational Linguistics (Coling 2010)}.

\bibitem[{Verga et~al.(2018)Verga, Strubell, and
  McCallum}]{verga-etal-2018-simultaneously}
Patrick Verga, Emma Strubell, and Andrew McCallum. 2018.
\newblock Simultaneously self-attending to all mentions for full-abstract
  biological relation extraction.
\newblock In \emph{Proceedings of the 2018 Conference of the North American
  Chapter of the Association for Computational Linguistics: Human Language
  Technologies, Volume 1 (Long Papers)}.

\bibitem[{Xu et~al.(2010)Xu, Stenner, Doan, Johnson, Waitman, and
  Denny}]{xu2010medex}
Hua Xu, Shane~P Stenner, Son Doan, Kevin~B Johnson, Lemuel~R Waitman, and
  Joshua~C Denny. 2010.
\newblock Medex: a medication information extraction system for clinical
  narratives.
\newblock \emph{Journal of the American Medical Informatics Association},
  17(1).

\bibitem[{Xu et~al.(2015{\natexlab{a}})Xu, Feng, Huang, and
  Zhao}]{xu-etal-2015-semantic}
Kun Xu, Yansong Feng, Songfang Huang, and Dongyan Zhao. 2015{\natexlab{a}}.
\newblock Semantic relation classification via convolutional neural networks
  with simple negative sampling.
\newblock In \emph{Proceedings of the 2015 Conference on Empirical Methods in
  Natural Language Processing}.

\bibitem[{Xu et~al.(2018)Xu, Wu, Wang, Yu, Chen, and
  Sheinin}]{xu-etal-2018-exploiting}
Kun Xu, Lingfei Wu, Zhiguo Wang, Mo~Yu, Liwei Chen, and Vadim Sheinin. 2018.
\newblock Exploiting rich syntactic information for semantic parsing with
  graph-to-sequence model.
\newblock In \emph{Proceedings of the 2018 Conference on Empirical Methods in
  Natural Language Processing}.

\bibitem[{Xu et~al.(2015{\natexlab{b}})Xu, Mou, Li, Chen, Peng, and
  Jin}]{xu-etal-2015-classifying}
Yan Xu, Lili Mou, Ge~Li, Yunchuan Chen, Hao Peng, and Zhi Jin.
  2015{\natexlab{b}}.
\newblock Classifying relations via long short term memory networks along
  shortest dependency paths.
\newblock In \emph{Proceedings of the 2015 Conference on Empirical Methods in
  Natural Language Processing}.

\bibitem[{Yin et~al.(2019)Yin, Song, Su, Zeng, Zhou, and Luo}]{yin2019graph}
Yongjing Yin, Linfeng Song, Jinsong Su, Jiali Zeng, Chulun Zhou, and Jiebo Luo.
  2019.
\newblock Graph-based neural sentence ordering.
\newblock In \emph{Proceedings of IJCAI}.

\bibitem[{Yu and Agichtein(2003)}]{yu2003extracting}
Hong Yu and Eugene Agichtein. 2003.
\newblock Extracting synonymous gene and protein terms from biological
  literature.
\newblock \emph{Bioinformatics}, 19.

\bibitem[{Zaremoodi and Haffari(2018)}]{zaremoodi-haffari-2018-incorporating}
Poorya Zaremoodi and Gholamreza Haffari. 2018.
\newblock Incorporating syntactic uncertainty in neural machine translation
  with a forest-to-sequence model.
\newblock In \emph{Proceedings of the 27th International Conference on
  Computational Linguistics}, pages 1421--1429.

\bibitem[{Zhang et~al.(2018{\natexlab{a}})Zhang, Liu, and
  Song}]{zhang-etal-2018-sentence}
Yue Zhang, Qi~Liu, and Linfeng Song. 2018{\natexlab{a}}.
\newblock Sentence-state lstm for text representation.
\newblock In \emph{Proceedings of the 56th Annual Meeting of the Association
  for Computational Linguistics (Volume 1: Long Papers)}, pages 317--327.

\bibitem[{Zhang et~al.(2018{\natexlab{b}})Zhang, Qi, and
  Manning}]{zhang-etal-2018-graph}
Yuhao Zhang, Peng Qi, and Christopher~D. Manning. 2018{\natexlab{b}}.
\newblock Graph convolution over pruned dependency trees improves relation
  extraction.
\newblock In \emph{Proceedings of the 2018 Conference on Empirical Methods in
  Natural Language Processing}.

\end{thebibliography}
\bibliographystyle{acl_natbib}

\end{document}